\RequirePackage{fix-cm}
\documentclass[twocolumn]{svjour3}          % twocolumn
\smartqed  % flush right qed marks, e.g. at end of proof
\usepackage{graphicx}
\begin{document}

%\title{Towards High Payload Capacity Latching Mechanism for Reconfigurable Robots  %\thanks{Grants or other notes
%\title{Latching Mechanism with High Payload Capacity for Reconfigurable Robots  %\thanks{Grants or other notes
%\title{LaMMos - Latching Mechanism based on Motorized-screw for Heavy Weight Reconfigurable Robots  %\thanks{Grants or other notes
\title{LaMMos - Latching Mechanism based on Motorized-screw for Reconfigurable Robots and Exoskeleton Suits %\thanks{Grants or other notes
}
%\subtitle{LaMMos - Latching Mechanism based on Motorized-screw}% for Reconfigurable Robots}


\author{Luis A. Mateos         \and
        Markus Vincze %etc.
}


\institute{Luis A. Mateos and Markus Vincze \at
              Gusshausstrasse 27 - 29 / E376, A - 1040, Austria. \\
              %Tel.: +43 1 58801 - 37601\\
              %Fax: +43 1 58801 - 37699\\
              \email{\{mateos, vinze\}@acin.tuwien.ac.at}           %  \\
}


%Category 6
% Category

\maketitle

\begin{abstract}

Reconfigurable robots refer to a category of robots that their components (individual joints and links) can be assembled in multiple configurations and geometries. Most of existing latching mechanisms are based on physical tools such as hooks, cages or magnets, which limit the payload capacity. Therefore, robots require a latching mechanism which can help to reconfigure itself without sacrificing the payload capability. 

This paper presents a latching mechanism based on the flexible screw attaching principle. In which, actuators are used to move the robot links and joints while connecting them with a motorized-screw and disconnecting them by unfastening the screw. 
The brackets used in our mechanism configuration helps to hold maximum force up to $5000N$. 
The LaMMos - Latching Mechanism based on Motorized- screw has been applied to the DeWaLoP - Developing Water Loss Prevention  in-pipe robot. It helps the robot to shrink its body to crawl into the pipe with minimum diameter, by reconfiguring the leg positions. And it helps to recover the legs positions to original status once the robot is inside the pipe. Also, LaMMos add stiffness to the robot legs by dynamically integrate them to the structure.

Additionally, we present an application of the LaMMos mechanism to exoskeleton suits, for easing the motors from the joints when carrying heavy weights for long periods of time.

This mechanism offers many interesting opportunities for robotics research in terms of functionality, payload and size.

\keywords{Reconfigurable Robot \and Latching Mechanism \and In-pipe  Robots \and Exoskeleton Suits}% \and Humanoid  Robots }
\end{abstract}

%%%%%%%%%%%%%%%%%%%%%%%%%%%%%%%%%%%%%%%%%%%%%%%%%%%%%%%%%%%%%%%%%%%%%%%%%%%%%%%%
\section{INTRODUCTION}

%Self-reconfigurable robots are autonomous kinematic machines with variable morphology. Beyond conventional actuation, sensing and control typically found in fixed-morphology robots, self-reconfiguring robots are also able to deliberately change their own shape by rearranging the connectivity of their parts, in order to adapt to new circumstances, perform new tasks, or recover from damage \cite{intro123123}.
%By saying "self-reconfigurable" it means that the mechanism or device is capable of utilizing its own system of control such as with actuators or stochastic means to change its overall structural shape \cite{4141950}.

   \begin{figure*}%[thpb]
      \centering
      \includegraphics[width=0.95 \textwidth]{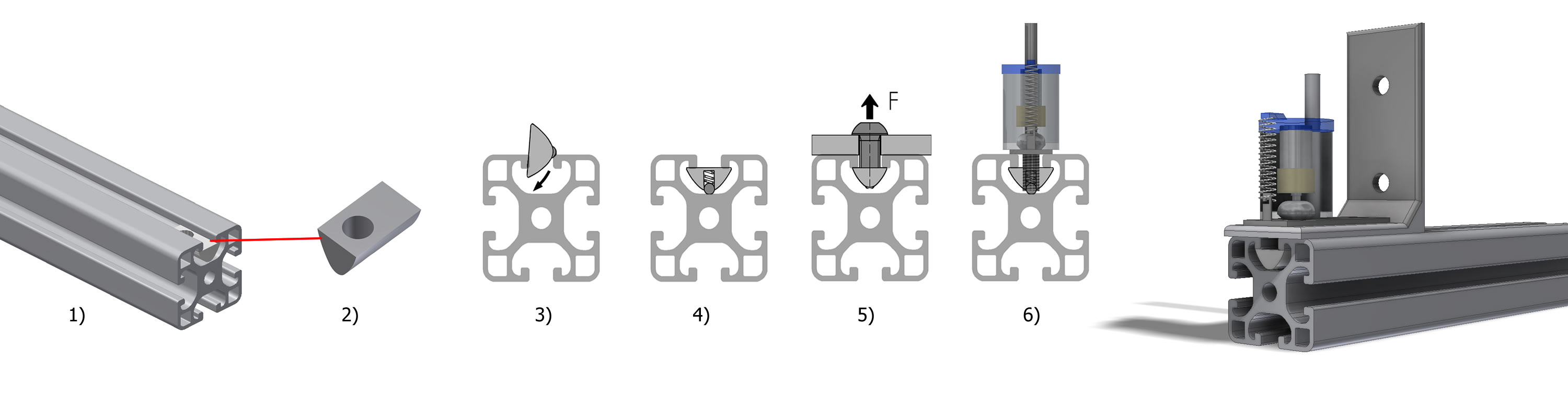}
      \caption{ Aluminum profiles representing the robot body where the LaMMos mechanism will be installed. 
      1) Aluminum profile with a special groove.
      2) T-Slot nut.
      3) The T-Slot nut is inserted into the profile groove.
      4) The T-Slot nut is set and trapped inside the profile groove.
      5) Bracket attached by driving the screw into the inserted T-Slot nut inside the profile. 
      6) LaMMos bracket attached to the aluminum profile with a motorized screw to the T-Slot nut. }
      \label{profiles}
   \end{figure*}

In order to be better adapted to various sized targets or complex geometric requirements, it is desirable that robots used in modern mechanical systems are geometrically reconfigurable. It means that the topological structure, kinematic parameters or dynamic parameters of the mechanism may be adjustable during the motion process \cite{5173816}.
A reconfigurable robot consists of a collection of individual links and joint components that can be assembled into multiple robot geometries. Compared to a conventional industrial robot with fixed geometry, such a system is able to provide flexibility, enabling itself to cope with a wide spectrum of tasks through proper selection and reconfiguration of a large inventory of functional components \cite{conro}.

Additionally, to the geometrical adjustment, in some cases, it is expected that the robot become stronger if it reconfigures its structure in order to carry loads beyond its initial capability or maintain its position passively and not actively consuming energy from the actuators.

Commonly, these reconfigurable mechanisms join the robot links with a latching mechanism, such as  hooks, cages or magnets \cite{crossball} \cite{Superbot} \cite{Miche} \cite{singo}. In this way, the mechanism is fast to attach and flexible to connect. %, however, lacks of payload capacity.
However, it has the limitation of restricting the payload capacity.
%In following sections we will present the existing latching mechanisms used in reconfigurable system. 
One can categorize them into two types, magnetic latching and physical latching.
\\
\\*
\textbf{Magnetic latching}\\*
Miche  (Modular Shape Formation by Self-Disassembly)  \cite{Miche} includes a connection mechanism by switchable magnets, able to connect to a neighborÕs steel plate and can support $2kg$. A similar latching mechanism is the M-TRAN (Self-Reconfigurable Modular Robotic System) \cite{mtran}, which is composed of nonlinear springs, Shape Memory Alloy (SMA) coils and magnets fixed on a moving part (connecting plate), able to lift two modules within the actual torque limit ($23kg$-$cm$).
As a result, the payload supported by these magnetics latching mechanism is relatively low if compared to physical latching connection mechanism.
\\
\\*
\textbf{Physical latching}\\*
A. Sproewitz \cite{4543747} presents a robust and heavy duty physical latching connection mechanism, which can ben seen as a hook with clamping principles. It can be actuated with DC motors to actively connect and disconnect modular robot units with load up to $18Kg$. 

Similar, the Superbot \cite{Superbot} module consists of six connectors, one on each side of the end effectors. Any of the six connectors of the Superbot module can connect to any connectors of another module with orientation intervals of 90$^\circ$.
The module's drivetrain for each degree-of-freedom (DOF) includes %The drive train of each degree of freedom of a module consists of 
a DC electric motor, a planetary gearbox, and an external gearbox, resulting in a maximum of $6.38Nm$ torque.
Given the size and weight of each module, this amount of torque is enough for reliably lifting three neighboring modules. 

JL-1 \cite{5173891} is a reconfigurable multi-robots system based on parallel and cone-shaped docking mechanisms. It is used for joining mobile robots to each other, in order to adopt a reconfigurable chain structure to cope with the cragged landforms which are difficult to overcome for a single robot.
Therefore, when two robots are linked, a full motorized spherical joint is formed.
This mechanism requires two motors on the docking side and one more motor on the driving platform connection. 

In contrast to the presented state of the art in latching mechanism, the LaMMos - Latching Mechanism based on Motorized-screw mechanism is able to support payloads up to $500kg$ and requires only one motor to make connection.

This paper describes the design and development of the LaMMos mechanism. 
Additionally, a couple of applications are presented, 
one for the DeWaLoP in-pipe robot and another for exoskeleton suits. 

For the DeWaLoP in-pipe robot, the LaMMos mechanism helps the robot to shrink its body to crawl into the pipe with minimum diameter, by reconfiguring the leg positions. And it helps to recover the legs positions as original once the robot is inside the pipe. Also, the LaMMos mechanism is used for increase the stiffness of the robot legs by dynamically integrate them to the structure. 

Another application of the LaMMos mechanism following the same principle of creating a rigid structure from movable joints is for exoskeleton suits \cite{4058569} \cite{6650376} \cite{4291584}. The LaMMos can improve the payload capacity of exoskeletons when these are required to carry heavy weights for long periods of time. %in the same position.

\section{Requirements for a latching mechanism in reconfigurable robot}

The aim of docking mechanism in reconfigurable robots is to attach/detach robot modules. 
%Several features a docking unit should be capable of can be distinguished 
There are a few requirements that a latching mechanism should fulfill \cite{intro} \cite{crossball} \cite{Superbot} \cite{Miche}.
However, the relevance of each single feature differs, depending on the functionality of the robot itself. 
Here we list the common requirements for a docking or connection mechanism used in self reconfiguration robots.

%In general, the following listed features are important for self-reconfigurable robots and should be provided by the bonding mechanism \cite{goodintro}:

   \begin{figure*}%[thpb]
      \centering
      \includegraphics[width=0.88 \textwidth]{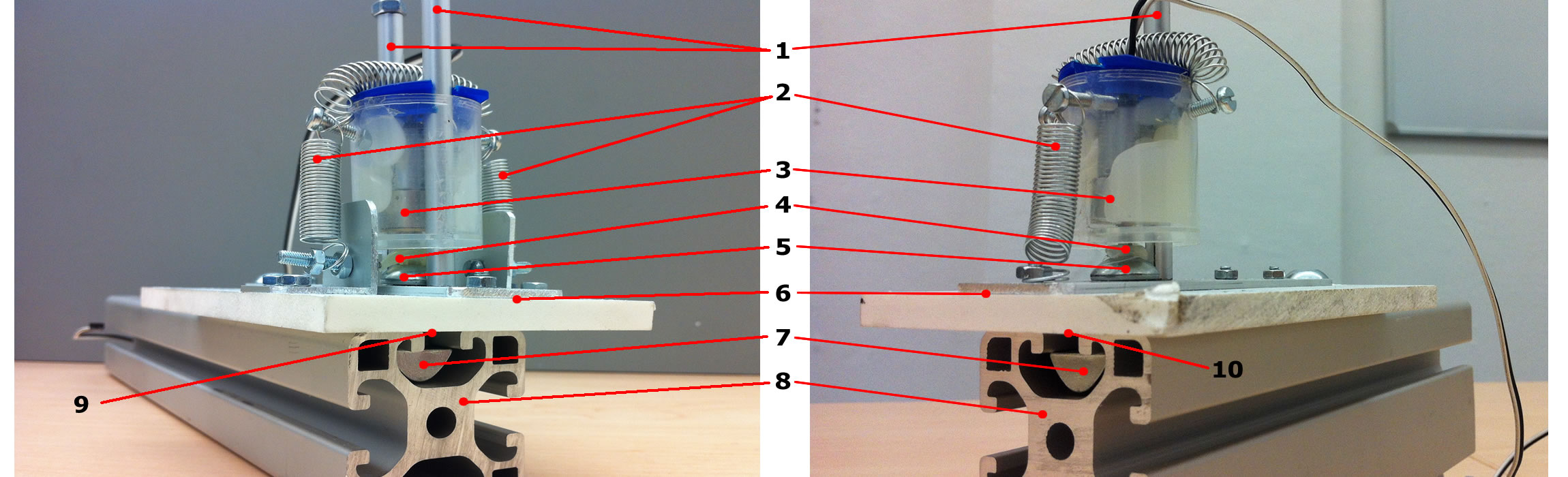}
      \caption{ Comparison of LaMMos standard (left) and simplified mechanisms (right). 
      1. Guiding system.
      2. Springs.
      3. Micro-geared motor. 
      4. Resin socket. 
      5. Screw.
      6. Bracket
      7. T-Slot nut.
      8. Aluminum profile.
      9. Moldable rubber inside the bracket.
      10. Flexible nut inside the bracket. }
      \label{hareconsistof15}
   \end{figure*}

\begin{description}
  \item[$\bullet$ Simple and fast docking procedure]
  \item[$\bullet$ Symmetric]
  \item[$\bullet$ Genderless]
  \item[$\bullet$ No accidental latching]
  \item[$\bullet$ Small size and durable]
  \item[$\bullet$ No power consumption in static state]
  \item[$\bullet$ Reliable power and signal transfer]
  \item[$\bullet$ Stable connection]
  \item[$\bullet$ Integration and protection of sensors]
  \item[$\bullet$ High latch load and impact strength]
  \item[$\bullet$ Few parts (especially moving ones)]
  \item[$\bullet$ Easy maintenance]
  \item[$\bullet$ Easy and low cost manufacturing and assembling]
\end{description}

The integration of all required features into a single functional mechanism is challenging and should be adapted to the purpose of the robot. %due to several constraints given by the robots design.
%For the development of LaMMos e.g. the no accidental latching, required a special mechanical property, which make it unique among lockable devices.
All mentioned features can be implemented into LaMMos mechanism except the genderless ability. This feature  is important for modular robots. Due that modules must be able to lock/unlock in any position. However, LaMMos is intended only for self-reconfigurable robot with predefined locking points.

%===========================================================
%===========================================================
\section{LaMMos Mechanism}

The LaMMos mechanisms enables robots to reconfigure its structure without loosing its payload capacity as other common latching mechanisms do. 
Also, the LaMMos mechanism enable robot joints to become rigid within its structure in order to handle heavy weight loads for long periods of time without loosing energy and protecting its movable actuators.

\begin{figure*}%[thpb]
      \centering
      \includegraphics[width=0.95 \textwidth]{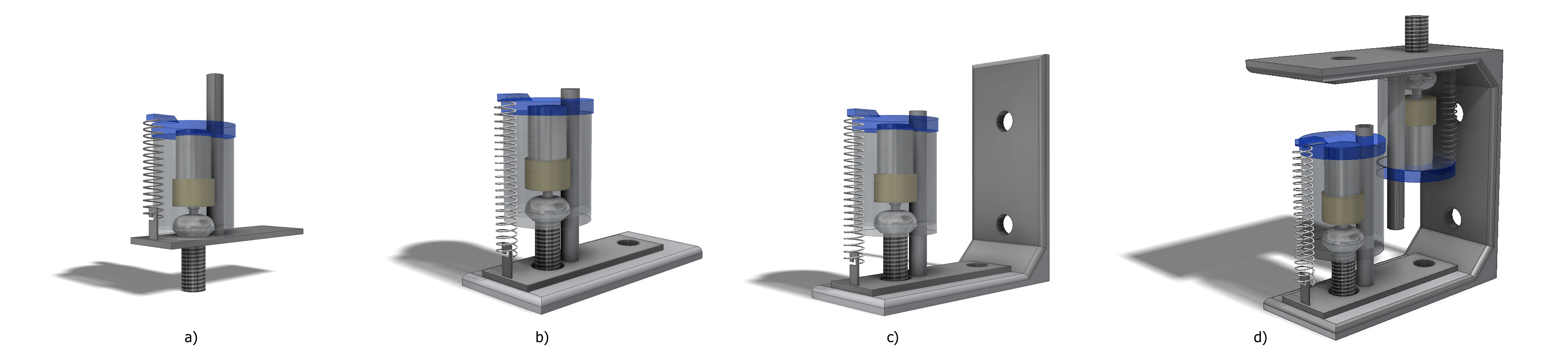}
      \caption{LaMMos mechanism integrated in brackets.
      a) LaMMos mechanism.
      b) LaMMos flat-bracket.
      c) LaMMos right-angle bracket.
      d) Multiple LaMMos inside a box-bracket.}
      \label{LaMMosTypes}
   \end{figure*}

The LaMMos mechanism adopts the flexible screw attaching principle for connecting or disconnecting robot parts.
It can be applied to any robots having rigid materialized surface, such as aluminum profiles, 
in which a T-Slot nut can be locked for further connection with brackets, as shown in Fig. \ref{profiles}. Moreover, LaMMos can be included in any type of bracket, such as flat, right angle, box, etc. Also, multiple LaMMos can be included in the same bracket, see Fig. \ref{LaMMosTypes}.

The LaMMos mechanism mimics the human operation of constructing robot links on the robot body by tightening a screw over a nut inside the body, and deconstructing the robot links by driving the screw out of the body.

   \begin{figure}%[thpb]
      \centering
      \includegraphics[width=0.47 \textwidth]{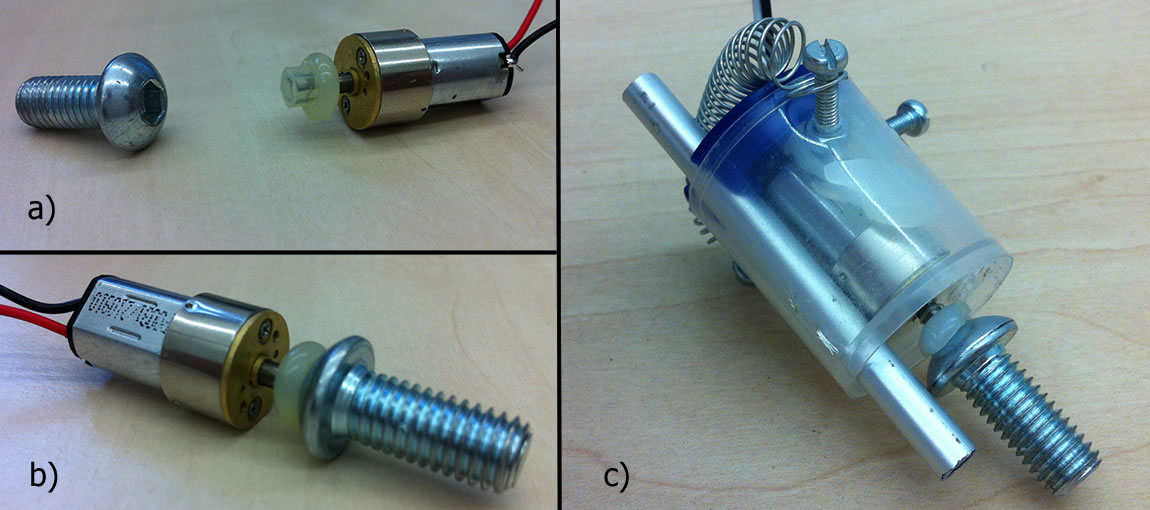}
      \caption{ LaMMos driving mechanism.
      a) Geared motor with resin socket for hexagonal head screw.
      b) Geared motor with screw integrated.
      c) Geared motor encapsulated with guiding tube.}
      \label{hareconsistof}
   \end{figure}

The main constituting elements of the LaMMos mechanism are divided into two main parts, the active part and passive part. The active part includes all the elements around the bracket: geared motor, screw, flexible nut, a compressed springs, a guiding tube %moldable rubber, couple of compressed springs, couple of guiding tubes 
and the bracket itself, as shown in Fig. \ref{hareconsistof15}. The passive part is the T-slot nut inserted into the robot body, or any device inside of the robot body that provide a nut for the screw.

The presented LaMMos mechanism is a simplified version of the standard LaMMos mechanism \cite{Ref:DeWaLoP_ICAR2013}. The main differences between these two versions are in the guiding mechanism. Since the motorized-screw is the same.

In the standard LaMMos version, the guiding mechanism includes a couple of compressed springs, two guiding rails and a moldable rubber inside the hole of the bracket. Whereas, in the simplified LaMMos, the guiding mechanism includes only one compressed spring, one guiding rail and a spring acting as a flexible nut located inside the hole of the bracket for guiding the motorized-screw.
\\ \\*
\textbf{T-Slot nut}\\*
The T-Slot nut is used for securing heavy components in fastening applications.
The T-Slot nuts are inserted into the profile groove where they are secured in position by driving a screw into it.
It will stay secure with holding up maximum force up to $5000N$, as shown in  Fig. \ref{profiles}.\\ \\*
\textbf{Driving mechanism - motorized-screw}\\*
The driving mechanism consists of a micro-geared motor with an integrated screw.
The geared motor specifications are in table \ref{motorspec}.

The screw is an hexagonal head $M8$ with a length of $18mm$.
A plastic resin is modeled to join the geared motor with the screw. It fits the motor shaft and the hexagonal screw head, as shown in Fig. \ref{hareconsistof}$a$ and Fig. \ref{hareconsistof}$b$.

The geared motor is encapsulated in a cylinder 
with radius of $27mm$ and length of height of $40mm$. The capsule includes a $6mm$ hole for the guiding system of the motorized-screw.

\begin{table}[thpb] 
\begin{center}
    \begin{tabular}{|l|l|l|}
    \hline
    Dimensions  & length = $33.2mm$  \\ 
    ~  & width = $14mm$  \\ 
    ~  & height = $14mm$   \\ \hline
    Gear ration  & 298:1   \\ \hline
    Stall Torque & $2884gm*cm$ at $3V$ \\ 
    ~ & $3444gm*cm$ at $6V$          \\ \hline
    Shaft        & $3mm$ diameter $D$ shaped         \\ \hline
    \end{tabular}
    \caption{Micro geared motor specifications.}
    \label{motorspec} 
    \end{center}
\end{table}
%\\ \\*
\noindent
\textbf{Guiding mechanism}\\*
The guiding mechanism of the simplified LaMMos consists of a single aluminum tube attached to the bracket, with diameter of $6mm$ and length of $70mm$. The tube crosses the cylinder where the geared motor is encapsulated, becoming its guiding rail in conjunction with the motorized-screw, as shown in Fig. \ref{hareconsistof}$c$.

In addition, the LaMMos mechanism includes a compressed spring with length of $19mm$ and width of $8mm$ with side hooks, see Table \ref{springsp}.
The function of the springs is to maintain the geared motor in touch with the bracket base when it tries to get out of it by rotating counterclockwise (unscrewing). In other words, the springs act as a pushing force for the motorized-screw to maintain its position when the screw gets loose.

\begin{figure*}%[thpb]
      \centering
      \includegraphics[width=0.95 \textwidth]{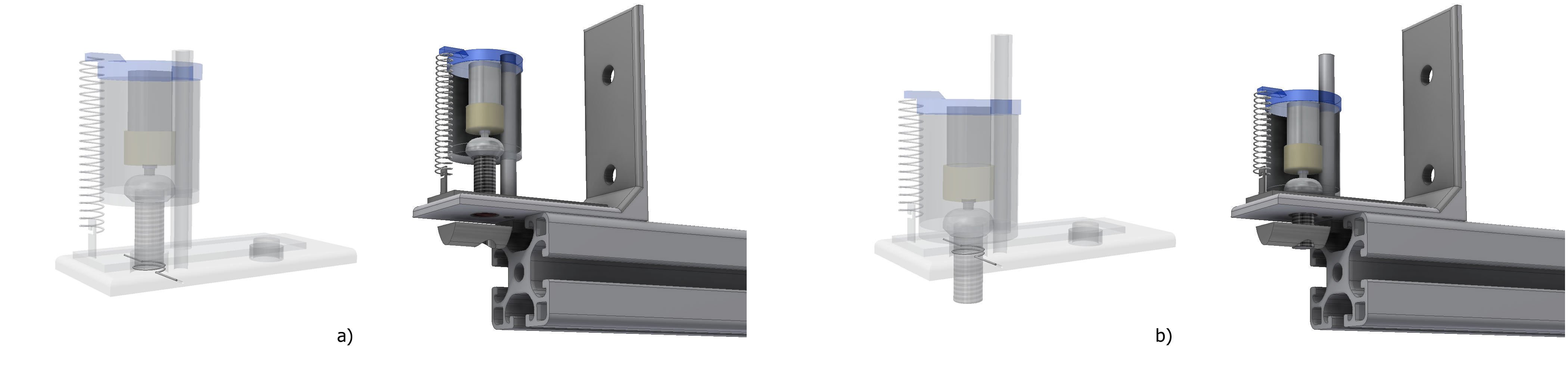}
      \caption{LaMMos mechanism in latching and unlatching status.
      a)  Unlatching status. The screw is above the bracket. It overcomes the force of the compressed spring by the support of the flexible nut.
      b) Latching status. The screw crosses the bracket. The flexible nut acts as initial thread for the screw to tight the T-Slot nut.}
      \label{haremodes}
   \end{figure*}

\begin{table}[thpb] 
\begin{center}
    \begin{tabular}{|l|l|l|}
    \hline
    Dimensions  & length = $19mm$  \\ 
    ~  & width = $8mm$  \\  \hline
    Diameter of spring wire, $d$  & $0.5mm$   \\ \hline
    Number of active coils, $n_a$  & $28mm$   \\ \hline
    Material & Stainless steel \\ \hline
    Weight  $M$ & $0.000406Kg$         \\ \hline
    Maximum load $F_{max}$ & $4.506N$         \\ \hline
    \end{tabular}
    \caption{Extension spring specifications.}
    \label{springsp} 
    \end{center}
\end{table}
%\\ \\*
\noindent
%There is an alternative configuration in which the compressed spring can be integrated in the LaMMos. It can be located at the top of the motorized-screw-capsule to a top roof linked to the guiding tube, as shown in Fig. \ref{alternative}. 
\textbf{Flexible nut}\\*
A flexible nut is located inside the hole of the bracket. 
Physically the flexible nut is similar to a torsion spring, it consists of a spring with one coil and with two opposite extension of the spring wires, which are inserted into the LaMMos bracket, as shown in figure \ref{haremodes}. 
The coil diameter is set to the diameter of the screw and the diameter of the wire is half millimeter diameter, so the thread of the screw is trapped. 
In the previous LaMMos mechanisms the flexible nut functionality was done by a moldable rubber. However, the moldable rubber can wear out over time. 
Since the functionality of the flexible nut is to guide the screw up or down from the bracket hole.

In this way, the flexible nut acts in two different ways: 
housing the screw and guiding the screw.\\ \\*
\textbf{Housing the screw}\\*
%\subsubsection{Housing the screw}
In order to move the LaMMos bracket, the screw must be housed in unlatching status, preventing any accidental latching. 
The screw is always pushed by the compressed springs towards the bracket, in order to maintain it over the bracket. With the friction provided by the flexible nut, the screw will not go straight through the bracket causing accidental latching, as shown in Fig. \ref{haremodes}$a$. In this mode, the LaMMos is set as a movable part.\\ \\*
\textbf{Guiding the screw to the T-Slot nut}\\*
%\subsubsection{Guiding the screw to the T-Slot nut}
During the latching process, the flexible nut provides an initial thread inside the bracket for the screw to go through until it reaches the T-slot nut inside the fastening target (robot body). The flexible nut has the property of guiding thread,  so that the screw will not be stopped and tight with it  before it  reaches the T-slot, as shown in Fig. \ref{haremodes}$b$.

Next, we introduce the DeWaLoP in-pipe robot and recall the multiple use of LaMMos mechanism in the robot.

%===========================================================
%===========================================================
%++++++++++++++++++++++++++++++++++++++++++++++++++++++++++++++++
%++++++++++++++++++++++++++++++++++++++++++++++++++++++++++++++++
\section{DeWaLoP In-PIPE ROBOT}

DeWaLoP stands for Developing Water Loss Prevention.
The goal of the DeWaLoP robot is to restore (repair, clean, etc,.) the over 100 years old pipe-joints of the fresh water supply systems of Vienna and Bratislava. 
These pipelines range from $800$ to $1000mm$ diameters and are still in good metallurgical shape. The pipe-joints have been detected as water loss points and therefore the DeWaLoP robot system is intended to crawl into these pipes and restore them \cite{Ref:MA31}. 

The DeWaLoP robot has large scale in size and weight. It has length of 1.4 meters and  radius of $380mm$, with weight from $200kg$.

\begin{figure*}%[b]
      \centering
      \includegraphics[width=0.88 \textwidth]{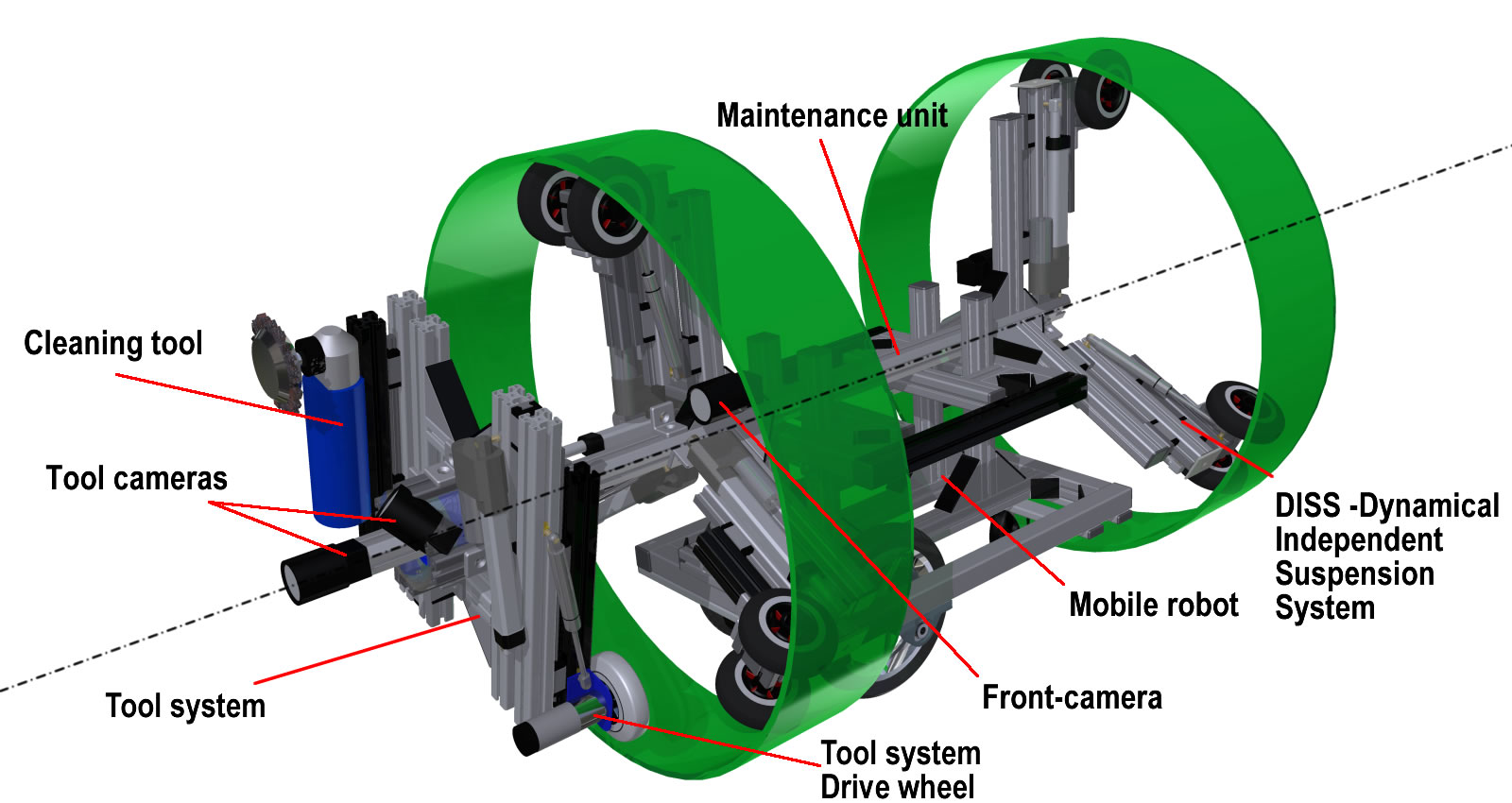}
      \caption{DeWaLoP in-pipe robot perspective view.}
      \label{dewalop}
   \end{figure*}

The robot consists of five main subsystems: control station, mobile robot, maintenance system, vision system and tool system, as shown in Fig. \ref{dewalop}:
\\ \\*
\textbf{Control station}\\*
%\subsection{Control station} 
The control station monitors and controls all the components of the in-pipe robot. The  controller includes a slate computer for monitoring and displaying the video images from the robot's Ethernet cameras.
Additionally, several 8 bits micro-controllers with Ethernet capabilities are included to send and receive commands to the in-pipe robot from the remote control joysticks and buttons \cite{Ref:DeWaLoP_ICAR20112}.
\\ \\*
\textbf{Mobile robot}\\*
%\subsection{Mobile robot} 
The mobile platform is able to move inside the pipes, carrying on board  electronic and mechanical components of the robot, such as motor drivers, power supplies, etc. It uses a differential wheel drive which enables the robot to promptly adjust its position to remain in the middle of the pipe while moving  \cite{Ref:DeWaLoP_ICMA2012}.
\\ \\*
\textbf{Maintenance unit}\\*
%\subsection{Maintenance unit}
The maintenance unit consists of a wheeled-leg structure able to extend or compress with a Dynamical Independent Suspension System (DISS) \cite{Ref:DeWaLoP_ICMET2011}. When extending its wheeled-legs, it creates a structure inside the pipe, so the robot tool work without  involuntary movements from its inertia. When compressing its wheeled-legs, the wheels become active and the maintenance unit is able to move along the pipe by the mobile robot.

The unit structure consists of six wheeled -legs, distributed in pairs of three, on each side, separated by an angle of 120$^{\circ}$, supporting the structure along the center of the pipe, as shown in Fig. \ref{dewalop}. The maintenance unit combines a wheel-drive-system with a wall-press-system, enabling the robot to operate in pipe diameters varying from $800mm$ to $1000mm$ \cite{Ref:DeWaLoP_IROS2013}. Moreover, the maintenance unit together with the mobile robot form a monolithic multi-module robot, which can be easily mounted/dismounted without the need of screws \cite{Ref:DeWaLoP_ICIRA2011}.
\\ \\*
\textbf{Vision system}\\*
%\subsection{Vision system}
The in-pipe robot includes four cameras, in order to navigate in the pipe, detect defects and redevelop specific areas \cite{Ref:DeWaLoP_CET2011}. 
\\ \\*
\textbf{Tool mechanism}\\*
%\subsection{Tool mechanism}
The tool mechanism enables the repairing of the pipe-joint in $3D$ cylindrical space \cite{Ref:DeWaLoP_ARW2012} \cite{Ref:DeWaLoP_ARW2013}.

%The concept of the DeWaLoP tool mechanism is based on the cylindrical robot principle, to cover $3D$ cylindrical space.
%However, the DeWaLoP mechanism modifies the standard cylindrical robot into a double cylindrical robot, where both arms are connected to the central axis and opposite each other.
%On one arm, the tool (cleaning or sealing) is integrated while on the other arm a drive-wheel rotates with high precision the entire tool mechanism \cite{Ref:DeWaLoP_ARW2012}. 

%++++++++++++++++++++++++++++++++++++++++++++++++++++++++++++++++
%++++++++++++++++++++++++++++++++++++++++++++++++++++++++++++++++
%===========================================================
%===========================================================

\section{Evaluation of the LaMMos Mechanism in DeWaLoP in-pipe Robot}
%\section{EXPERIMENTAL STUDY CASE - DeWaLoP}

\begin{figure}[b]
      \centering
      \includegraphics[width=0.47 \textwidth]{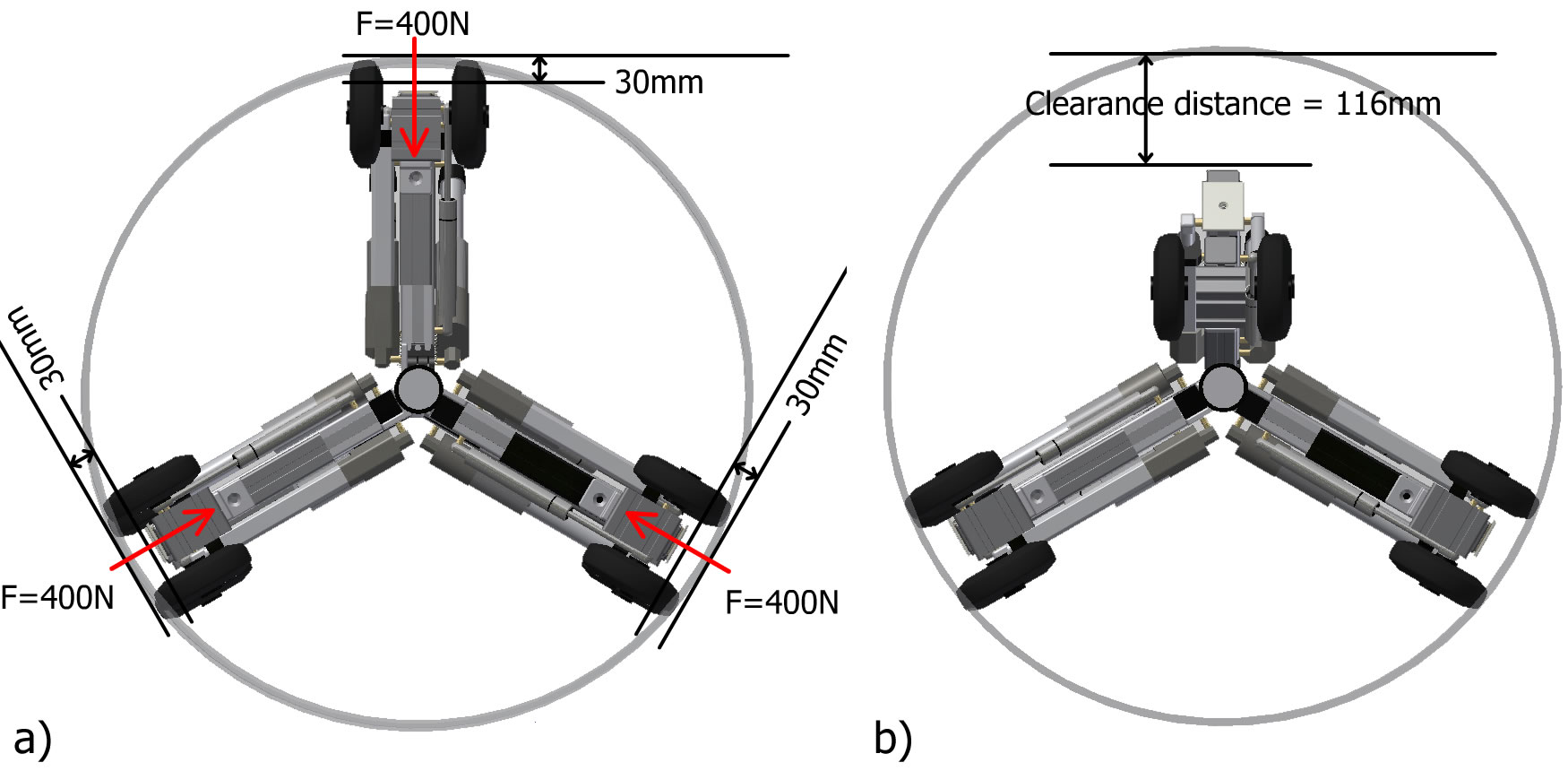}
      \caption{ DeWaLoP in-pipe robot front view. 
      a) Before installing the LaMMos mechanism, to insert the robot into the $800mm$ diameter pipe. All wheeled-legs must compress its $400N$ gas spring by $30mm$ for clearance of the legs to the pipe.
      b) Inserting the robot into the $800mm$ diameter pipe with the top wheeled-legs lowered 
      by LaMMos mechanism, enabling a clearance distance of $116mm$ for easy insertion and protection of the wheeled-legs from hitting the pipe while entering.
}
      \label{xxxx}
   \end{figure}

The LaMMos mechanism is intended to improve the DeWaLoP robot in two different ways. The first is to reconfigure the top wheeled-legs for easy input of the robot in the smaller pipe diameter. And the second is to add stiffness to the wheeled-legs once they are extended forming a centered structure inside the pipe. \\ \\*
\textbf{Easy robot insertion}\\*
The DeWaLoP robot has been designed to work in pipes with diameters ranging from $800mm$ to $1000mm$, where the robot is able to move and perform the redevelopment task. 
However, to insert the robot into the $800mm$ diameter pipe requires precision and effort, as the gas springs of each wheeled-leg must be compressed by $30mm$ for creating the space to enter the pipe, as shown in Fig. \ref{xxxx}$a$. The compress force needs to be at least $400N$. %effort. 
%This is because all wheeled-legs must compress its $400N$ gas spring by $30mm$ for clearance of the legs to the pipe, as shown in Fig. \ref{xxxx}. 
In other words, the operators must push the robot into the pipe and at the same time push each of its wheeled-legs with a force $F\geq400N$, to compress the springs and insert the robot into the $800mm$ diameter pipe.

Instead of using brute force to insert the robot into the pipe, an alternative solution is to lower its top wheeled-legs. 
In such way, the robot can be easily inserted into the pipe with minimal effort, protecting the legs from hitting the pipe while entering it, as shown in Fig. \ref{xxxx}$b$. 
And once inside the pipe the robot reconfigure the wheeled-legs as original.
\\ \\*
\textbf{Stiffness to extended wheeled-legs}\\*
Another constraint of the in-pipe robot is that once inside the pipe and in locations where its required to rehabilitate the pipe. The wheeled-legs extend creating a centered structure. However, the maximum force the legs can hold is limited to linear actuator specifications. 

Hence, if a locking mechanism attaches each wheeled-leg to the maintenance unit structure, then the dynamical structure formed by extending the legs increases its stiffness.

%Hence, if once the wheeled-legs are extended, a locking mechanism attaches the wheeled-leg to the maintenance unit, then the dynamical structure form be extending the legs becomes a static one with higher rigidity.

%Its rigidity is not optimal, since the stiffness is given by the wheeled-legs actuators capacity. %Consequently, this can be improved if a 

\begin{figure} %[!t]
      %\centering
      \includegraphics[width=0.48 \textwidth]{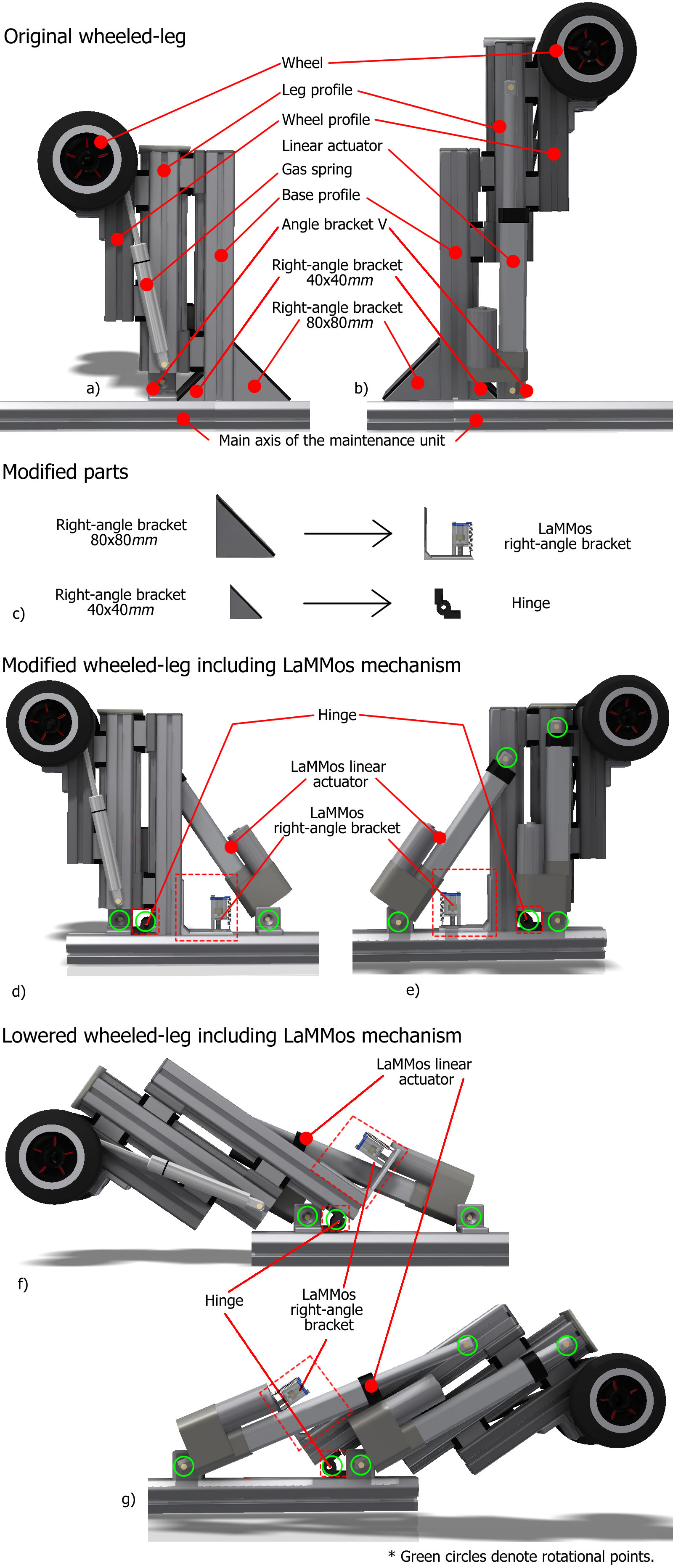}
      \caption{a) Wheeled-leg (left side view), revealing the gas spring connected from the $wheel$ profile to the $leg$ profile. 
      b) Wheeled-leg (right side view), revealing the linear actuator for extending/contracting the leg. 
      c) Modified parts: The $80 \times 80mm$ right-angle bracket is replaced by a LaMMos right-angle bracket. The $40 \times 40mm$ right-angle bracket is replaced by a hinge. % to become the rotational support of the $base$ profile.
      d) Wheeled-leg modified with LaMMos mechanism installed (left side view). 
      e) Wheeled-leg modified (right side view).
      f) Lowered wheeled-leg  with LaMMos bracket housing its motorized-screw (left side view).
      g) Lowered wheeled-leg (right side view).
      %* The green circles mark the rotational point.
      }
      \label{wheeledlegs}
   \end{figure}

In following sections we will first describe the structure of the robot wheeled-legs,  then how we install the two proposed LaMMos mechanism into the legs. 
Finally, we will present the procedure of our experiment showing how the LaMMos mechanism helps to reconfigure the robot and make it stronger.\\ \\*
\textbf{Wheeled-leg before installing LaMMos}\\*
A wheeled-leg in DeWaLop robot consists of the following components, see Fig. \ref{wheeledlegs}.

1) $Base$ profile. 

2) $Leg$ profile.

3) $Wheel$ profile.

4) Linear actuator.

5) Gas spring.

6) Angle bracket $80\times80mm$.

7) Angle bracket $40\times40mm$.

8) Angle bracket $V$.

\begin{figure}%[thpb]
      \centering
      \includegraphics[width=0.48 \textwidth]{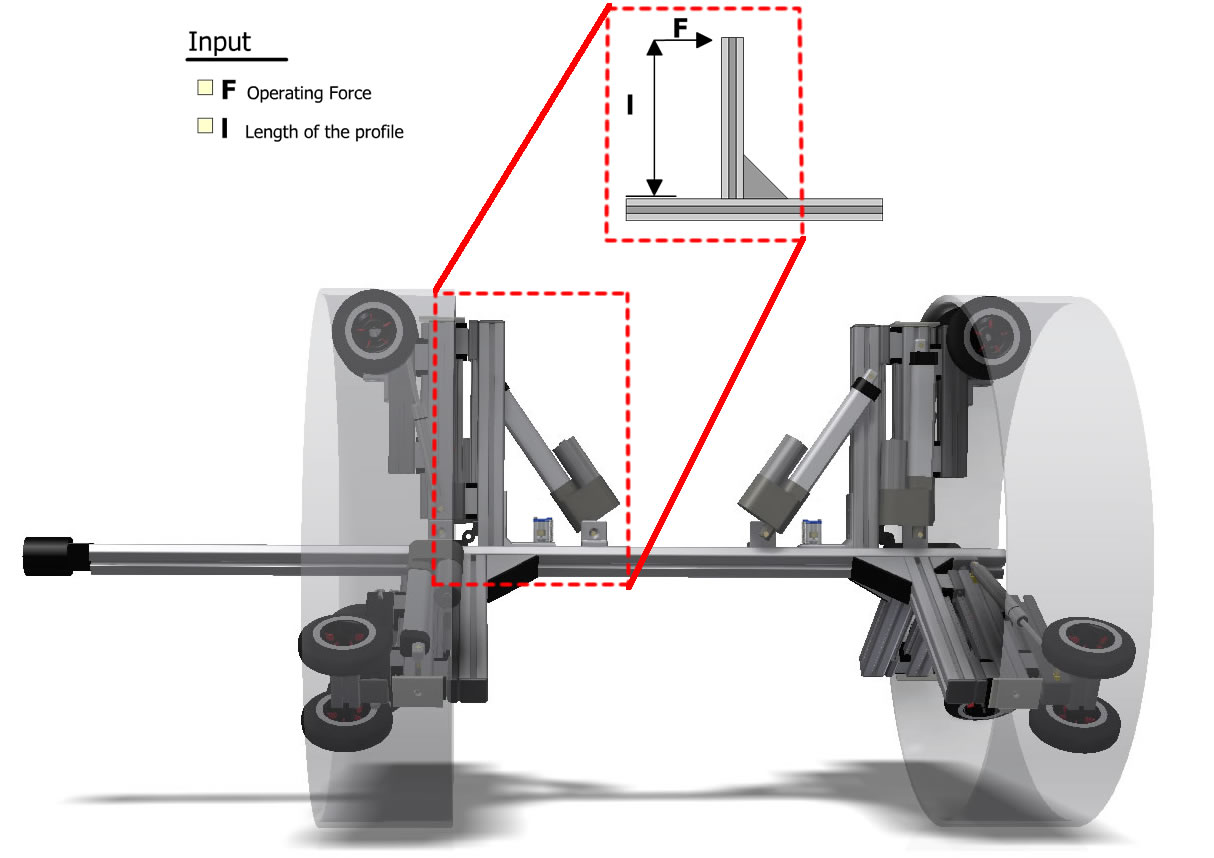}
      \caption{Wheeled-leg load - carry capacity of the right angle bracket.}
      \label{wheeledlegsload}
   \end{figure}

The right-angle brackets are characterized by its high load-bearing capacity to overcome displacement, torsion and deflection.
As shown in Table \ref{tab:anglefastenersforce}, where $F$ refers to the operating force that the bracket can hold, and $l$ refers to the length of the corresponding part attached by the bracket, see Fig. \ref{wheeledlegsload}.
%refers to the length of the corresponding part hold by the bracket.

\begin{table}[ht] 
\begin{center}
    \begin{tabular}{|l|l|}
    \hline
     Bracket 8  $40\times40mm$  & $F<1000N \wedge F \times l < 50Nm$    \\ \hline
     Bracket 8  $80\times80mm$  & $F<2000N \wedge F \times l < 150Nm$    \\ \hline
     Bracket 8  $160\times80mm$  & $F<2000N \wedge F \times l < 150Nm$    \\ \hline
    \end{tabular}
    \caption{Load - carry capacity of the right angle bracket.}
    %Right angle bracket load-carry capacity from applied Force $F$ and Length $l$. }
    \label{tab:anglefastenersforce} 
    \end{center}
\end{table}

%The wheeled-leg configuration is as follows:
%The $base$ profile supports all wheeled-leg components. 
The $base$ profile is the main support for the wheeled-leg components as it is attached with a $80\times80mm$ and a $40\times40mm$ right-angle brackets to the maintenance unit axis, as shown in Fig. \ref{wheeledlegs}$a$ and \ref{wheeledlegs}$b$. %right-angle brackets to the maintenance unit main axis. 
%Opposite to it, a bracket $40\times40mm$ helps supporting the $base$ profile to the maintenance unit, as shown in Fig. \ref{wheeledlegs}$a$ and \ref{wheeledlegs}$b$. 
To the $base$ profile face with the $40\times40mm$ right-angle bracket, linear rails are installed to match the linear bearings from the $leg$ profile. 
In this way, the $leg$ can be extended with a linear actuator, which is attached to the $leg$ and to the maintenance unit with an angle bracket $V$. In this configuration, the linear actuator extends the leg by pushing and contracts the leg by pulling, as shown in Fig. \ref{wheeledlegs}$b$.
On the $leg$ profile, parallel and opposite to the $base$ profile, another linear rail with bearings is installed, attaching the $wheel$ profile enabling it to move up or down. %The movements of the $wheel$ are given by the suspension system (gas spring).  
Additionally, the $wheel$ profile is supported by an extended gas $spring$, connecting the $wheel$ with the $leg$, acting as a suspension system, as shown in Fig. \ref{wheeledlegs}$a$.

%\\ \\*
\noindent \\*
\textbf{Wheeled-leg after installing LaMMos mechanism for easy robot insertion}\\*
The LaMMos mechanism helps DeWaLoP robot to adjust its wheeled-legs to a lower height position before entering the pipe and afterwards helps to recover the original vertical position once the robot is sitting inside the pipe.
The right-angle bracket $80\times80mm$ is replaced by the LaMMos right-angle bracket, while the $40\times40mm$ bracket is substituted by a hinge, as shown in Fig. \ref{wheeledlegs}$c$.

In this configuration, the functionality of LaMMos is to attach/detach the $80\times80mm$ right-angle bracket from the $base$ profile to the maintenance unit. 
The functionality of the hinge is to keep the wheeled-leg in contact with the maintenance unit when the LaMMos is detached. % while the wheeled-leg is lowered.  
In other words, the hinge is required as a joint rotational connection between the $base$ profile and the maintenance unit.

Additionally, for lowering the wheeled-leg, a LaMMos linear actuator is required. % to move it in 1D. 
 The linear actuator pushes the leg to be lowered, as shown in Fig. \ref{wheeledlegs}$f$ and \ref{wheeledlegs}$g$. And by pulling it, sets the wheeled-leg to original position which is perpendicular to the maintenance unit. see Fig. \ref{wheeledlegs}$d$ and \ref{wheeledlegs}$e$.
%So, when pushing, the wheeled-leg is lowered, as shown in Fig. \ref{wheeledlegs}$d$. While, when pulling, the wheeled-leg is set as original, perpendicular to the maintenance unit, see Fig. \ref{wheeledlegs}$c$.

%In this way, the LaMMos mechanism acts as the original $80\times80mm$ right-angle bracket from the $base$ profile, fastening the wheeled-leg to the main axis of the maintenance unit. 
In the stage when the leg is vertical, the LaMMos mechanism acts as the replaced $80\times80mm$ right-angle bracket from the $base$ profile, as it fastens the wheeled-leg to the maintenance unit using a motorized-screw.
It is able to overcome heavy payload due to the rigid structure of LaMMos bracket. %the maximum payload forces from the bracket itself, see table \ref{tab:anglefastenersforce}. 

   \begin{figure}%[thpb]
      \centering
      \includegraphics[width=0.48 \textwidth]{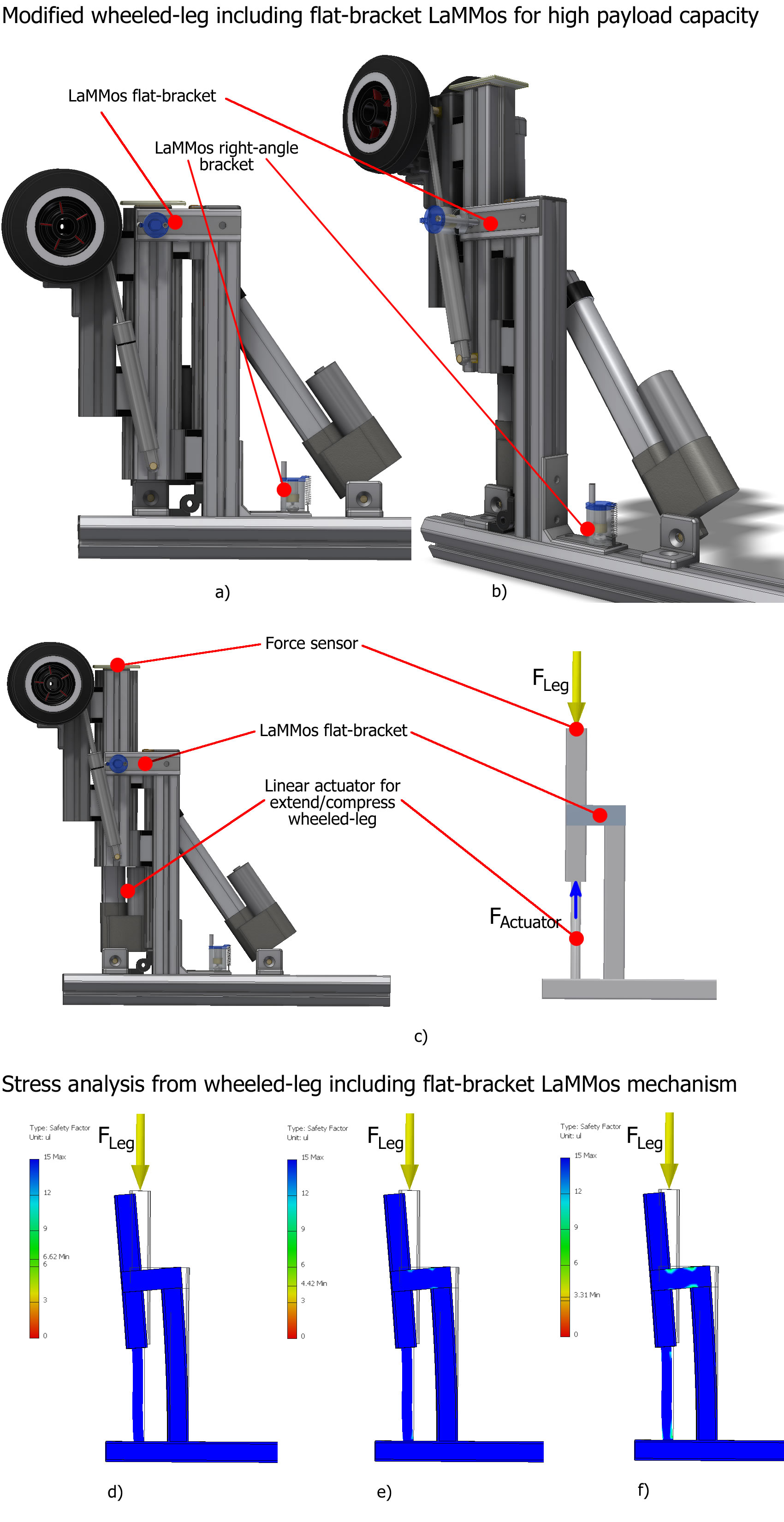}
      \caption{a) Compressed wheeled-leg with integrated flat-bracket LaMMos mechanism in unlatched state.%for high payload capacity. 
      b) Extended wheeled-leg with integrated flat-bracket LaMMos mechanism in unlatched state.
      c) LaMMos mechanism latching the $leg$ to the $base$ profile and its simplified model.
      d) Stress analysis test for force load $F_{Leg}=1000N$.
      e) Force load of $F_{Leg}=1500N$.
      f) Force load of $F_{Leg}=2000N$.
}
      \label{wheeledlegs2x}
   \end{figure}

If the leg hit obstacles when performing restoration task, Table \ref{tab:anglefastenersforce} shows the maximum forces $F$ that brackets with various parameters are able to hold.
We are using $80\times80mm$ bracket for our LaMMos mechanism, so its payload capacity is up to $2000N$.\\ \\*
\textbf{Wheeled-leg after installing LaMMos mechanism for high payload capacity}\\*
Each wheeled-leg includes a linear actuator for extend or compress the leg in conjunction with a linear slide. In this configuration, when the wheeled-leg is extended, the points of contact from the $leg$ to the $base$ profile are a couple of linear bearings and a linear actuator, as shown in Fig. \ref{wheeledlegs}$b$.
The forces acting from the "foot" of the $leg$ $F_{Leg}$ to the linear actuator cannot be higher than the maximum load capacity  of the actuator $F_{Actuator}$, otherwise, it will be damaged.

  \begin{figure*} [t]
      \centering
      \includegraphics[width=0.98 \textwidth]{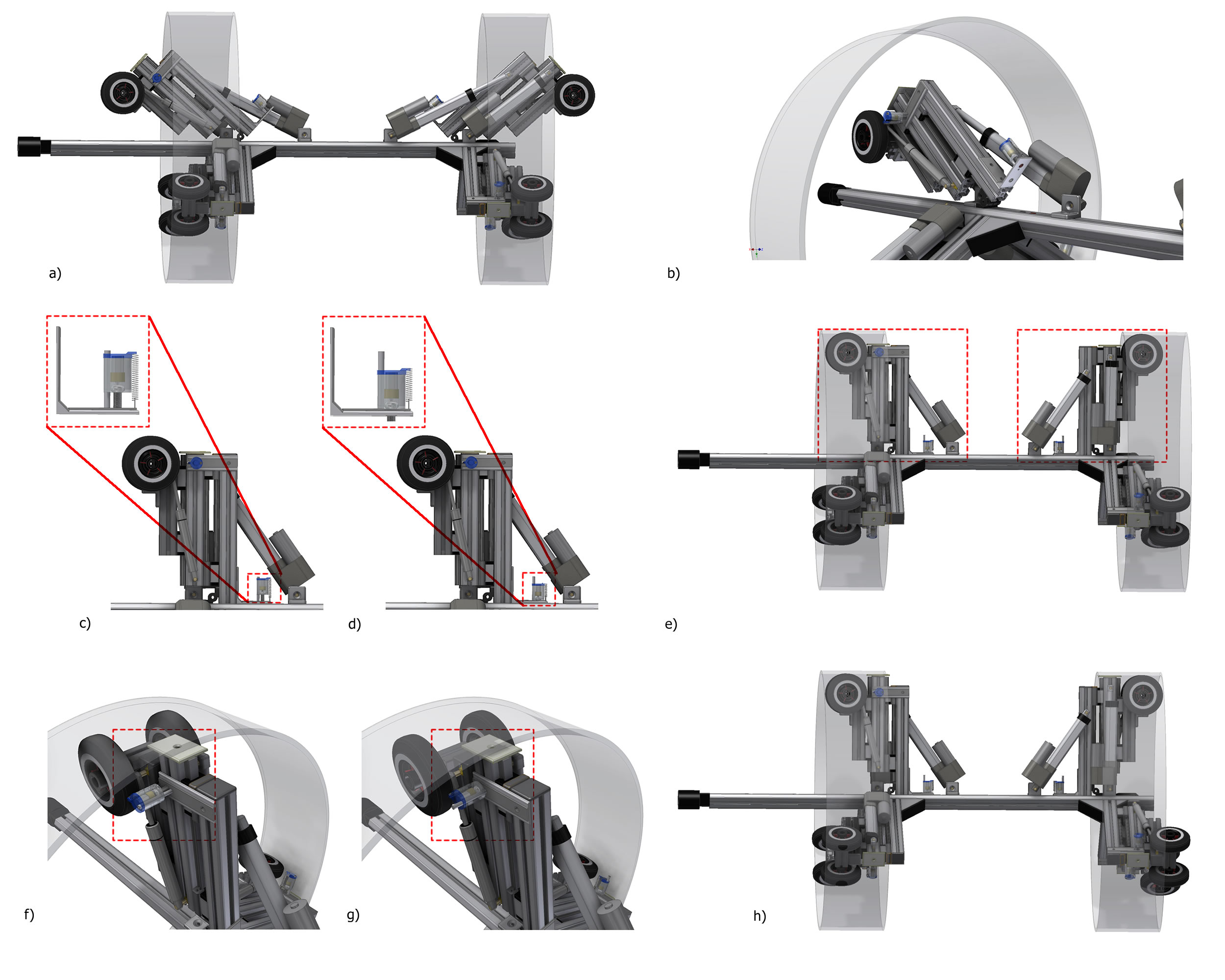}%3dprocess.png}
      \caption{ %LaMMos mechanism installed in DeWaLoP top wheeled-legs.
      a) DeWaLoP robot with top wheeled-legs lowered for easy insertion to pipes with $800mm$ diameter.
      b) LaMMos mechanism $housing$ the motorized-screw. 
      c) Wheeled-leg retracted with LaMMos mechanism $housing$ the motorized-screw.
      d) Wheeled-leg retracted with LaMMos mechanism fasten to the maintenance unit. The motorized-screw is tight to the T-Slot nut inside the maintenance unit profile.
      e) DeWaLoP robot with top wheeled-legs re-constructed as original due to the LaMMos mechanism and able to overcome the maximum payload forces from the bracket itself.
      f) $Leg$ extended with LaMMos flat-bracket mechanism $housing$ the motorized-screw.
      g) LaMMos flat-bracket mechanism fastening the $leg$ to the $base$ profile of the maintenance unit. The motorized-screw is tight to the T-Slot nut inside the $leg$ profile.
      h) DeWaLoP robot with extended wheeled-legs forming a rigid structure inside the pipe. Since, the wheeled-legs become part of the robot structure.
      }
      \label{3dprocess}
   \end{figure*}

In order to increase dynamically the load capacity of the maintenance unit. 
The wheeled-legs must be modified to include another LaMMos mechanism as a flat-bracket connecting the $leg$ to the $base$ profile, as shown in  Fig. \ref{wheeledlegs2x}.

When the legs are compressed, the LaMMos flat-bracket is unlatched. On the other hand, when the legs are extended and positioning the robot in the center of the pipe, the LaMMos flat-bracket is latched.
The functionality of the LaMMos is to dynamically connect the legs to the structure of the maintenance unit, in order to add stiffness to the structure and protect the linear actuators.
In this way, the forces acting on the $leg$ profile $F_{Leg}$ will be damped by the structure of robot and not directly by the linear actuators. \\ \\*
\textbf{Safety factor}\\*
Consequently, the wheeled-legs are able to hold forces beyond its actuator load capacity by including the LaMMos mechanism. 
The included linear actuator in each wheeled-leg is able to hold a maximum load of $1000N$. 
In Fig. \ref{wheeledlegs2x}$d,e,f$, the safety factors are obtained from simulated loads of $1000N$, $1500N$ and $2000N$ acting directly on the extended leg. For this simulation the LaMMos flat-bracket is a steel plate of $l=116m$, $h=40mm$ and $w=9mm$.

%In this configuration, when the robot extend its wheeled legs to become a rigid centered structure inside the pipe, the LaMMos mechanism lock the legs to the maintenance unit structure,as the legs become part of the stiff structure. In this way, the structure is able to damp forces beyond its movable actuators, while protecting them.

%In this configuration, when the robot becomes a rigid centered structure inside the pipe after extending its wheeled legs to the inner pipe surface, the LaMMos mechanism lock the legs to the maintenance unit structure,as the legs become part of the stiff structure. In this way, the structure is able to damp forces beyond its movable actuators, while protecting them.

%The LaMMos flat-bracket is attached securely to the $base$ profile with the ability to latch or unlatch to the $leg$ profile dynamically.
The safety factor \textbf{SF} is a term describing the structural capacity of a system beyond the expected loads. Factor of safety guidelines include the following:

$\rightarrow$ A safety factor less than 1.0 at a location indicates that the material at that location has failed.

$\rightarrow$ A safety factor larger than 1.0 at a location indicates that the material at that location is safe.

$\rightarrow$ For many applications, a \textbf{SF} of 4 is a common goal, especially if product durability is an issue.
\\ \\*
\textbf{DeWaLoP robot reconfiguration process}\\*
The insertion and reconfiguration process of the DeWaLoP robot inside the pipe is as follows:

%with the lowered wheeled-legs 
%reconfiguration process to insert the DeWaLoP robot with the lowered top wheeled-legs and then inside the pipe reconstruct the legs as original with the LaMMos mechanism
%is as follows:

{\bf Step 1.} Initially the LaMMos linear actuator pushes the top wheeled-legs of the DeWaLoP robot, to enter into the pipe with diameter of $800mm$, as shown in Fig. \ref{3dprocess}$a$.
At this point the wheeled-legs are not rigid, as its only points of contacts to the maintenance unit are the LaMMos linear actuators, the hinges and the linear actuators for extending/compressing the leg. 

%Initially the wheeled-leg is constructed, the LaMMos right-angle bracket is tightening the $base$ profile to the maintenance unit main axis. At this point, the wheeled-leg is able to overcome forces up to $2000N$, as shown in Fig. \ref{3dprocess}$b$. 

%The angle bracket is screw to the T-Slot nut and the maintenance unit is constructed to overcome the maximum force calculated, due to its structure and not due to its linking mechanism.

{\bf Step 2.} With the operation from the robot remote control, the robot moves into the pipe as it is positioned inside the pipe, see Fig. \ref{3dprocess}$b$. The linear actuators pull the wheeled-legs until the legs are perpendicular to the maintenance unit axis, see Fig. \ref{3dprocess}$c$. At this point the LaMMos right-angle brackets are at the right position but is not tight to the T-Slot nut inside the profile.

{\bf Step 3.} Then, the LaMMos mechanism activates the motorized-screw to rotate clockwise until the screw tights the T-Slot nut inside the maintenance unit, as shown in Fig. \ref{3dprocess}$d$. 

%The LaMMos mechanism rotates the screw counterclockwise until housing the screw, as shown in Fig. \ref{3dprocess}$c$. 

{\bf Step 4.} The wheeled-legs are re-constructed, the LaMMos right-angle bracket has fasten the $base$ profile to the maintenance unit axis. Each of the wheeled-leg is able to overcome forces up to $2000N$, as shown in Fig. \ref{3dprocess}$e$.

{\bf Step 5.} The robot is located inside the pipe and required to rehabilitate a pipe-joint. It extend all its wheeled-legs and once a centered structure has been reached, the LaMMos flat-bracket (integrated on all the wheeled-legs) are activated, connecting the $legs$ to the $base$ profiles of the maintenance unit. In this way, the $legs$ become part of the maintenance unit structure, as shown in Fig. \ref{3dprocess}$g$. 
Resulting in a rigid structure able to overcome higher forces that its movable actuators, as shown in Fig. \ref{3dprocess}$h$. 
%Resulting in a rigid structure able to overcome forces up to $5000N$, as shown in Fig. \ref{3dprocess}$f$. 
\\ \\*
\textbf{Evaluation of LaMMos mechanism}\\*
In the process of attaching the LaMMos bracket to the T-Slot nut inside the maintenance unit, the motorized-screw requires a power supply of $3V$ delivering a stall torque of $2884gm*cm$, which is enough to tight the bracket securely.
While for unscrewing it, the supplied voltage of the motorized-screw is doubled ($6V$) with a stall torque of  $3444gm*cm$.%, so the mechanism is able to always unscrew itself. 

  \begin{figure}[thpb]
      \centering
      \includegraphics[width=0.47 \textwidth]{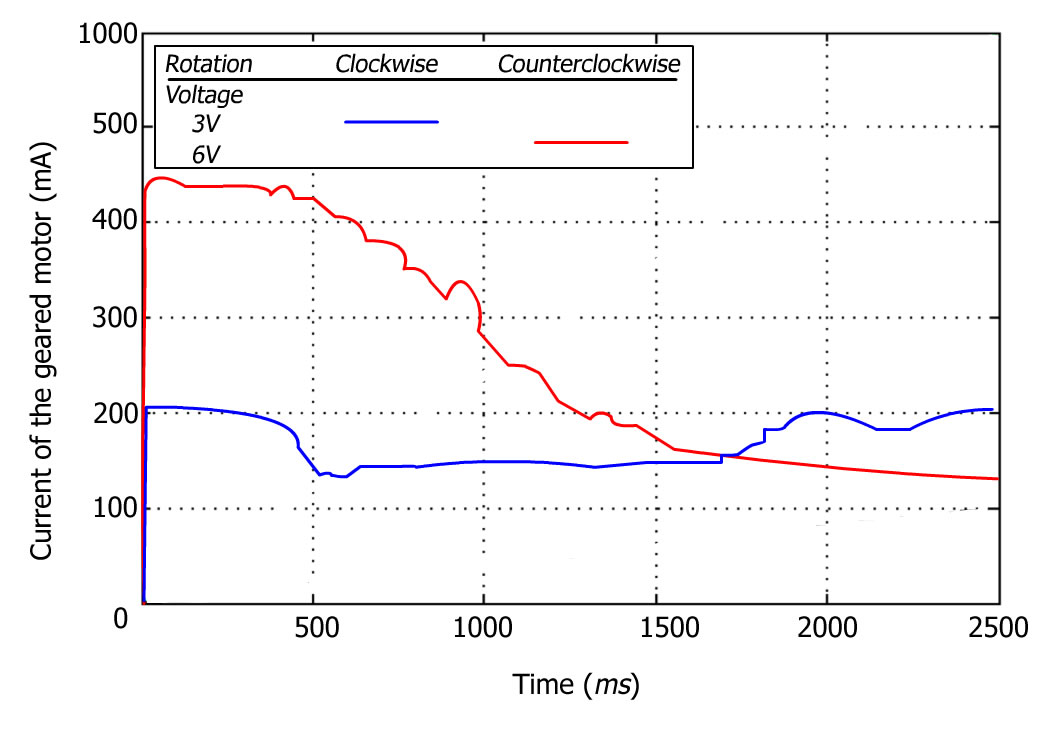}
      \caption{ Current (mA) / Time (t) plot from the LaMMos's geared motor. }
      \label{3dproceasdasdssx}
   \end{figure}

As shown in Fig. \ref{3dproceasdasdssx}, initially the LaMMos is housing the screw, then the geared motor starts to rotate clockwise at voltage of $3V$. The screw starts to follow the thread from the flexible nut, crossing it and finally reaching the T-Slot nut.
To unscrew the motorized-screw, the voltage supply is doubled to $6V$. The geared motor reaches a peak current while starting to rotate counterclockwise. Once the screw is out of the T-Slot nut, the current stabilizes while the screw follows the thread of the flexible nut until it is housed inside the LaMMos bracket.

   \begin{figure}[b]
      \centering
      \includegraphics[width=0.47 \textwidth]{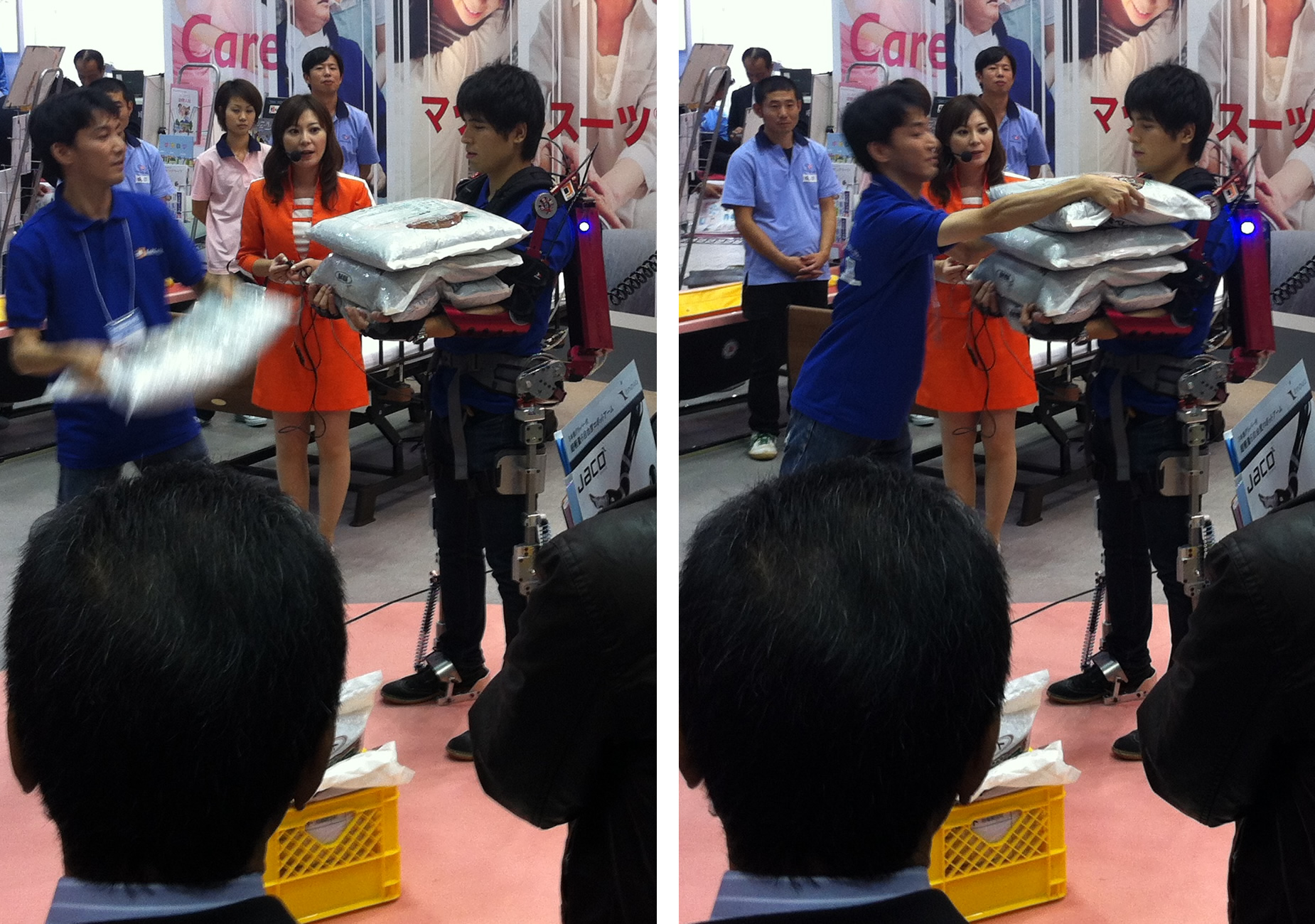}
      \caption{Electric exoskeleton payload capacity demonstration.}
      \label{exo}
   \end{figure}

\section{LaMMos for Exoskeleton Applications}

An exoskeleton is an external structural mechanism with joints and links corresponding to those of the human body \cite{4291584}. 
There are two main groups of exoskeletons, the ones with unlimited power supply including both wearable types with a tether or those fixed to a base \cite{1639189} \cite{5975512}.  And the ones that carry their own power supply  \cite{4108030} \cite{1491470}.

   \begin{figure*}[t]
      \centering
      \includegraphics[width=0.95 \textwidth]{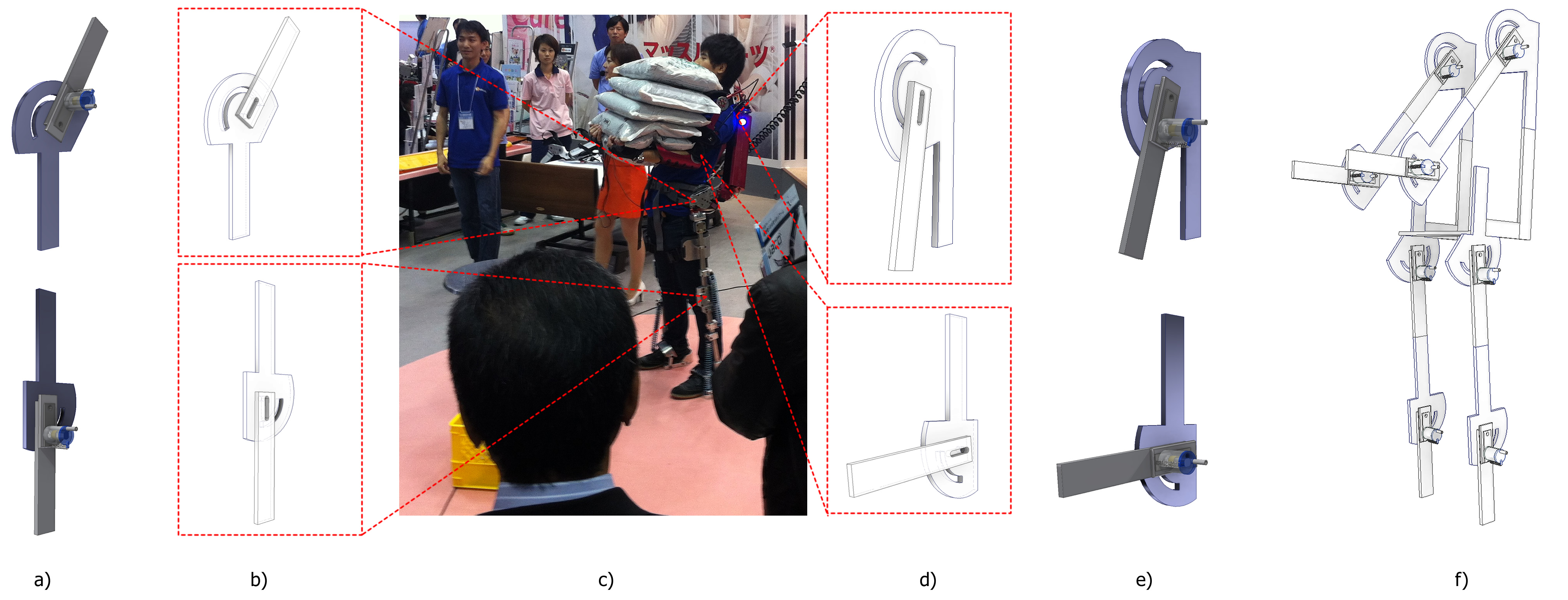}
      \caption{ 
      a) Lower joints including LaMMos mechanism.
      b) Lower joints guiding slots.
      c) Electric exoskeleton payload capacity demonstration.
      d) Upper joints guiding slots.
      e) Upper joints including LaMMos mechanism.
      f) Proposed exoskeleton including LaMMos mechanism for high payload capacity on critical joints.}
      \label{exo2}
   \end{figure*}

Exoskeletons with unlimited power supply are able to power its actuators and motors for unlimited time. On the other hand, exoskeletons with limited power supply must avoid positions that have high consume of energy in order to save energy and operate longer. % and last longer. 
Such positions include carrying heavy weights with extended arms, as shown in Fig. \ref{exo}. 
As a result, the available power impose strong limitations on a battery powered exoskeleton \cite{5979863}. 
\\ \\*
\textbf{The HAL robot suit}\\*
The current HAL (Hybrid Assistive Limb) suit, HAL-5, is  a full body exoskeleton that carries its own power supply. It  consists of frames interconnected by power units that each contain an electric motor and reduction gears and are positioned directly next to the hip, knee, shoulder (flexion) and elbow joints of the wearer to assist his movements  \cite{4108030}.

Additional passive DoF are located at each shoulder, upper  arm, and ankle joint. The suit is powered by batteries. The system is controlled according to the intentions of the wearer, which are obtained by measuring the bioelectric signal (BES) on the skin above the main flexor and extensor 
muscles associated with each augmented human joint. Motor torques are calculated according to these signals. 
\\ \\*
\textbf{Exoskeleton joints}\\*
The mechanisms included in exoskeleton joints usually combine slots and rollers mimicking the rolling and sliding of human bones \cite{5979761}. 
Also, these exoskeleton joints are limited in strength and in flexibility due to its mechanical configurations and elements \cite{4058569}.

For exoskeletons with limited power supply, servomotors are efficient by including permanent magnets with the capacity of stepping-down gearing to provide high torque and responsive movement in a small package \cite{5975494}  \cite{6491207}. Geared servomotors can also utilize electronic braking to hold in a steady position while consuming minimal power.
However, even with the most efficient servomotor there will be losses of energy. 
%Consequently, the use of a locking mechanism inside the joints will help to maintain the exoskeleton position stiff without using power, increasing the battery life substantially.
\\ \\*
\textbf{LaMMos mechanism in exoskeleton joints}\\*
Suppose the classic demonstration of exoskeleton, in which the wearer is required to carry heavy weights with extended arms for a long period of time, see Fig. \ref{exo2}$c$. Or suppose the case of an exoskeleton in the emergency of an earthquake, in which it is holding the roof of a building enabling an exit for people trapped inside. 
In these case scenarios, the battery of the exoskeleton will drain fast, due to the demanding power by the motors. 

  \begin{figure*} [t]
      \centering
      \includegraphics[width=0.95 \textwidth]{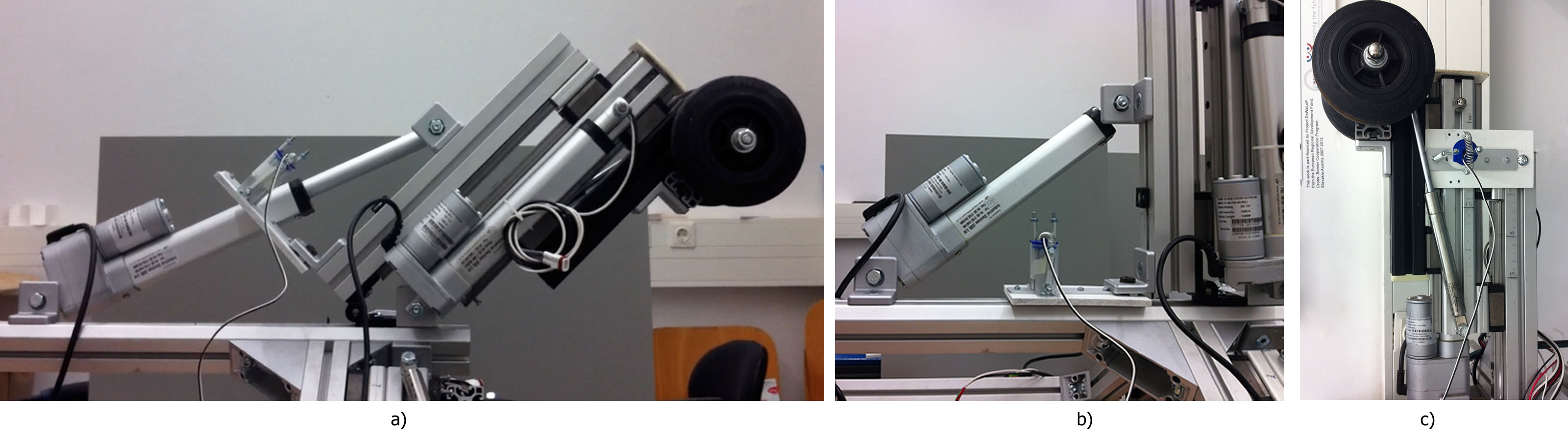}
      \caption{Photos of LaMMos mechanism on DeWaLoP robot. 
      a) The LaMMos linear actuator is pushing the wheeled-leg to lower its position, while the LaMMos bracket is housing the motorized-screw. 
      b) LaMMos mechanism in active mode, fastening the leg to the maintenance unit. 
      c) LaMMos flat-bracket for high payload capacity attaching the $leg$ to the $base$ profile of the wheeled-leg. 
}
      \label{3dprocesazzzzasx}
   \end{figure*}

Therefore, we propose to add to the exoskeleton joints the LaMMos mechanism, in order to lock the position of the limbs dynamically when an increase of power consumption is detected or when required by the user. 
From Fig. \ref{exo2}$b$ and $d$, it is possible to notice that the lower and upper joints mobility are given by the guiding slots. Hence, if the movable plate of the joint integrates a flat-bracket LaMMos for high payload capacity, as shown in Fig. \ref{exo2}$a$ and $e$. 
Then, the limb can be locked securely in any position and the servomotors from the joints set to standby. And when the task is finished and the exoskeleton is required to move again, the LaMMos are deactivated enabling movement to the exoskeleton as normal. %and the servo motors from the joints activated and able to move as normal. 

In this configuration, the joints by including a LaMMos mechanism for high payload capacity are able to dynamically reconfigure the exoskeleton structure to a static rigid structure at any time, see Fig. \ref{exo2}$f$.

%Thus, the exoskeleton joints will be able to lock its position dynamically when an increase of power consumption is detected or whenever required, becoming a rigid exoskeleton, saving the batteries energy for more useful tasks.

%Hence, if the LaMMos mechanism is included in this exoskeleton, at a specific position it can become a rigid exoskeleton, saving the batteries for more useful tasks.

%From Fig. \ref{exo2}, the operator inside the exoskeleton is required to hold several sacks for long periods of time. Then would useful to lock the position of the arms and knees  with a latching mechanism, without using the motors from the joint to hold the weight. 

%The structural design and materials have for objective to be as light and
%Its structure consists of hard-lightweight materials, such as aluminum, titanium, carbon fiber, etc. 
%In order to reduce the weight of the structure and reduce the amount of torque and energy needed from the actuators when moved.
%Ideally, an exoskeleton should be as light as possible, 

%There has been for passive gravity compensation in exoskeletons significantly reduces the amount of torque and energy needed from the actuators 

%
%============================
%============================
%

\section{CONCLUSIONS}

This paper introduces a Latching Mechanism based on Motorized-screw (LaMMos) for heavy weight reconfigurable robot. This mechanism improves the payload that other common mechanisms provide.

%The LaMMos mechanism implies screw attachment principle of all the fastening elements. In this way, the latching mechanism consists of a bracket with special features, including a motorized-screw. %And a passive T-Slot nut  located on the destination point.

The LaMMos mechanism requires one motor per each screw, and one actuator per each moving dimension. %For moving in 3D space its requires 3 actuators, for moving in 2D space requires 2 and for one dimension only one actuator.
Hence, the LaMMos mechanism is able to move the robot components in one, two or three dimensional space. And
attach the components to the robot body or to other robot parts.
%the other robot parts.%, by firmly tight its bracket with a motorized-screw into a specific part of the robot.%, to a T-Slot nut. 

The LaMMos mechanism can be integrated into any type of bracket, we presented the LaMMos as right-angle bracket and as flat-bracket. Moreover, several LaMMos can be included in a single bracket.

The presented LaMMos mechanism is useful when the reconfigurable parts of the robot are required to maintain payloads beyond the limit of its movable actuator.
It is able to reconstruct a robot, as if a person with a screwdriver tights the screw on the nut. 
In this way the robot assemble is optimal as the components are latched with each other as if they are constructed.

For the LaMMos mechanism evaluation, the DeWaLoP in-pipe robot was used and modified, as shown in Fig. \ref{3dprocesazzzzasx}. 
Initially a couple of its wheeled-legs were modified to include the right-angle LaMMos bracket instead of rigid brackets. The objective is to lower the wheeled-legs so the in-pipe robot is able to enter its minimal working pipe diameter while protecting it legs. And once inside the pipe reconstruct the robot wheeled-legs as original.
The acting forces on the legs of the robot may reach $2000N$ and therefore a latching mechanism with strong stability is required.

%Moreover, all wheeled-legs are modified to include a flat-bracket LaMMos mechanism for high payload capacity. The aim of this LaMMos is to add stiffness to the centered structure formed from extending the wheeled-legs once the robot is inside the pipe and ready to rehabilitate the pipe-joint. In this way, the wheeled-legs from been dynamical parts of the robot, become rigid with the structure, by latching each leg to the robot structure. Thus, the robot is able to support forces beyond the payload capacity of its actuators.

%Otherwise, with dynamically "soft" re-configurable mechanisms the wheeled-legs may break or permanently damage its joints. 

%Thus, this novel mechanism offers many interesting opportunities for robotics research in terms of functionality, payload and size. Since we can reconfigure the robot by constructing it.

\addtolength{\textheight}{-5cm}   % This command serves to balance the column lengths

%%%%%%%%%%%%%%%%%%%%%%%%%%%%%%%%%%%%%%%%%%%%%%%%%%%%%%%%%%%%%%%%%%%%%%%%%%%%%%%%

%%%%%%%%%%%%%%%%%%%%%%%%%%%%%%%%%%%%%%%%%%%%%%%%%%%%%%%%%%%%%%%%%%%%%%%%%%%%%%%%

%%%%%%%%%%%%%%%%%%%%%%%%%%%%%%%%%%%%%%%%%%%%%%%%%%%%%%%%%%%%%%%%%%%%%%%%%%%%%%%%

\section*{ACKNOWLEDGMENT}
This work is part-financed by Project DeWaLoP from the European Regional Development Fund, Cross- Border Cooperation Programme Slovakia- Austria 2007-2013.

%%%%%%%%%%%%%%%%%%%%%%%%%%%%%%%%%%%%%%%%%%%%%%%%%%%%%%%%%%%%%%%%%%%%%%%%%%%%%%%%

%\begin{thebibliography}{99}
%\end{thebibliography}

%\bibliographystyle{IEEEtran}
\bibliographystyle{plain}
%\bibliography{references}

\end{document}